\documentclass[letterpaper, 10 pt, conference]{ieeeconf}  

\IEEEoverridecommandlockouts                              

\usepackage{graphics} 
\usepackage{epsfig} 
\usepackage{amsmath} 
\usepackage{amssymb}  

\usepackage{tikz}

\usepackage{multirow}
\usepackage{booktabs}
\usepackage{url}

\usepackage{rotating}
\usepackage{subfig}

\definecolor{orange}{rgb}{1,0.5,0}

\newcommand{\vx}{\mathbf{x}}

\newcommand{\best}[1]{\underline{\textbf{#1}}~}

\newcommand{\R}{{\rm I\!R}}

\title{\LARGE \bf
Benign-Malignant Lung Nodule Classification with
Geometric and Appearance Histogram Features
}

\author{Tizita Nesibu Shewaye$^{1}$, Alhayat Ali Mekonnen$^{2}$
\thanks{$^{1}$MedTech Graduate, University of Applied Sciences - Wiener Neustadt}%
\thanks{$^{2}$LAAS-CNRS, Univ de Toulouse, LAAS, France}%
}

\begin{document}

\maketitle
\thispagestyle{empty}
\pagestyle{empty}

\begin{abstract}
Lung cancer accounts for the highest number of cancer deaths globally. 
Early diagnosis of lung nodules is very important to reduce the mortality rate 
of patients by improving the diagnosis and treatment of lung cancer. 
This work proposes an automated system to classify lung nodules as malignant and benign in CT images.
It presents extensive experimental results using a combination of geometric
and histogram lung nodule image features and different linear and non-linear
discriminant classifiers. The proposed approach is experimentally validated
on the LIDC-IDRI public lung cancer screening thoracic computed tomography (CT) dataset
containing nodule level diagnostic data. The obtained results are very encouraging
correctly classifying $82\%$ of malignant and $93\%$ of benign nodules
on unseen test data at best.
\end{abstract}

\section{INTRODUCTION} \label{sec:introduction}

Uncontrolled abnormal cell growth in any part of the body leads to space occupation which might be a cancer. 
When these abnormal cells are growing in the lung area they might lead to lung cancer. 
Lung cancer accounted for $19.5\%$ of the total cancer related deaths in 2012 -- the highest number of all 
cancer related deaths globally~\cite{Fer13}. Lung cancer appears as
pulmonary nodules which are small round or oval-shaped growth in the lung. But, all
pulmonary nodules are not cancerous and in fact over $90\%$ of pulmonary nodules that are
smaller than two centimeters in diameter are benign complicating proper diagnosis.

The main problem with lung cancer is that the majority of patients have evidence of spread at
the time of presentation~\cite{EMI13}. But, early diagnosis can improve the effectiveness of
treatment and increase the patient's chance of survival; hence, early detection through
screening is of vital importance. Of the utilized imaging modalities to screen for lung cancer, it
has recently been shown that Computed Tomography (CT) screening does actually lead to
reduced deaths from lung cancer~\cite{Abe11}. Consequently, radiologists will have to screen several
scans on a daily basis. This puts increased burden which could lead to mistakes due to the
overwhelming number of cases handled. To alleviate this burden, Computer Aided Diagnosis
(CADx) systems can be used to help radiologists in terms of both accuracy and speed. Some
studies have indeed shown improvements in radiologist's performance through the use of CAD
systems, e.g.,~\cite{Rub05}. In line with this, this work investigates image processing and
machine learning techniques for automated lung nodule benign and malignant classification.

This work focuses on benign and malignant nodule classification, primarily for two reasons: 
(i) it is the least investigated category in the literature~\cite{ElB13}, 
and (ii) it has been recently getting more attention to a point that there
are grand challenges organized on it~\cite{Armato_2015}. 
Since the malignancy of lung nodules correlates highly with their geometrical size, shape, and appearance,
this work proposes to investigate pattern recognition and machine learning techniques to
automatically classify benign and malignant nodules based on these features. Specifically, the
study focuses on evaluation of different linear and non-linear discriminant classifiers 
extensively used in the machine learning and pattern recognition domains -- namely: logistic
regression, linear Support Vector Machines (SVM), K-nearest neighbors (K-NN),
discrete AdaBoost, and random forest -- with a heterogeneous 
feature set componsed of geometric, gray scale histogram, and oriented gradient histogram features
extracted from CT images. The proposed approach is experimentally validated
on the LIDC-IDRI public lung cancer screening thoracic computed tomography (CT) dataset
containing nodule level diagnostic data. The obtained results are very encouraging,
correctly classifying $82\%$ of malignant and $93\%$ of benign nodules
on unseen test data at best.

\subsection{Related Work}


The main approach utilized in the literature for lung nodule classification follows
 a two step approach which uses a feature extraction and classification steps
\cite{kumar2015lung,Li2005357}. 
In these approaches, the classifiers are trained using labeled dataset in a supervised
manner. Unfortunately, since most of them report experimental results using
their own proprietary dataset that is not publicly available 
or a different subset of a publicly available dataset, a direct absolute comparison
of their performance is not possible. Nevertheless, pertinent works are summarized in Table~\ref{tbl:soa:feat_summ}.

\begin{table}[!ht]
\resizebox{0.49\textwidth}{!}{
\begin{tabular}{ccccccc}
\toprule
\multirow{2}{*}{Work}	&	\multicolumn{4}{c}{Image Features}			&	Clinical Features	&	Classifier	\\ \cmidrule(r){2-5}
						&	Geometric	&	Appearance	&	Texture		&	2D/3D 	&				\\
\toprule
Way et al.~\cite{Ted_Way_2009}& 	$\checkmark$ &	$\checkmark$ & $\checkmark$	&	3D		&	&	LDA, SVM	\\
Way et al.~\cite{Ted_way_2006}&		$\checkmark$ &	$\checkmark$ & $\checkmark$	&	3D		&	&	LDA		\\
Armato et al.~\cite{armato_2003}&	$\checkmark$ &  $\checkmark$	&			&	3D		&	&	LDA	\\
Lee et al~\cite{Lee201043}		& $\checkmark$   &  $\checkmark$	&   $\checkmark$	&	3D 	&	$\checkmark$  & 	LDA \\
Tartar et al.~\cite{Tartar_2014}& $\checkmark$   &					&					&	2D	& $\checkmark$  		& Ensemble Classifiers\\ 
Aoyama et al.~\cite{Aoy02}		&				 & 	$\checkmark$ 	&			&  	2D	&	$\checkmark$ &	ANN	\\
Li et al.~\cite{Li2005357}		&				 & 	$\checkmark$ 	& 	$\checkmark$ 	&	2D	&		&   LDA	\\
Orozco et al.~\cite{MaderoOrozco2015}&			 &					&	$\checkmark$  	&	2D	&		&	SVM \\
Kumar et al.~\cite{kumar2015lung}\footnotemark	 
								&			 & 	$\checkmark$ 	& 	$\checkmark$ 	&	2D	&		&	ANN, Decision Tree	\\			
\bottomrule		
\end{tabular}
}
\caption{Summary of relevant work in benign-malignant lung nodule classification.}
\label{tbl:soa:feat_summ}
\end{table}

\begin{figure*}[!ht]  
\centering
\begin{tikzpicture}[every text node part/.style={align=left},xscale=1.75,yscale=1.75]
\small
\node[text width=2.5cm,align=center] at (1.0,2.85) {Nodule ROI};
  \draw [line width=0.35mm,->] (1.5,2.25) -- (2.125,2.25);
  \draw [line width=0.35mm,->] (4.5,2.25) -- (5.125,2.25);
  \node[inner sep=0pt] (nodule) at (0.9,2.25) {\includegraphics[width=.1\textwidth]{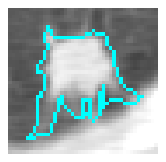}};
  \node[inner sep=0pt] (nodule) at (3.9,2.25) {\includegraphics[width=.14\textwidth]{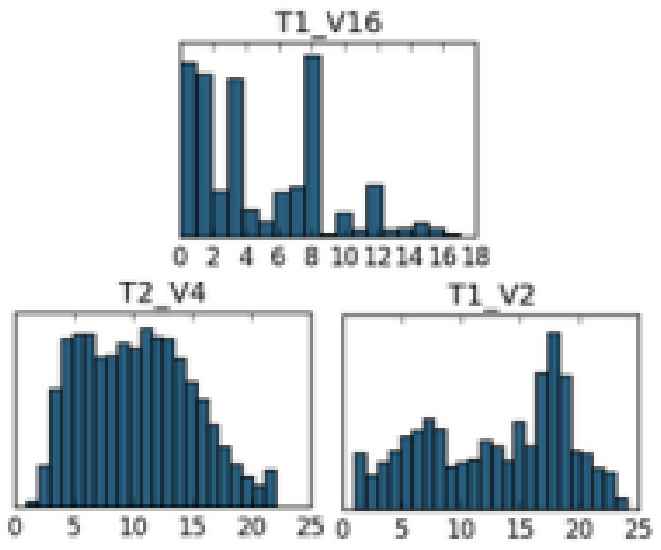}};
  \draw[rounded corners=1ex] (2.125,1.75) rectangle (4.5,2.75) node[pos=.5,text width=3.5cm] {Feature \\ Extraction};
  \draw[rounded corners=1ex] (5.125,1.75) rectangle (7.5,2.75) node[pos=.5,text width=3.5cm] {Trained \\ Classifier};
  \draw [line width=0.35mm,->] (7.5,2.25) -- (8.125,2.25);

  \node[inner sep=0pt] (classifier) at (6.9,2.25) {\includegraphics[width=.12\textwidth]{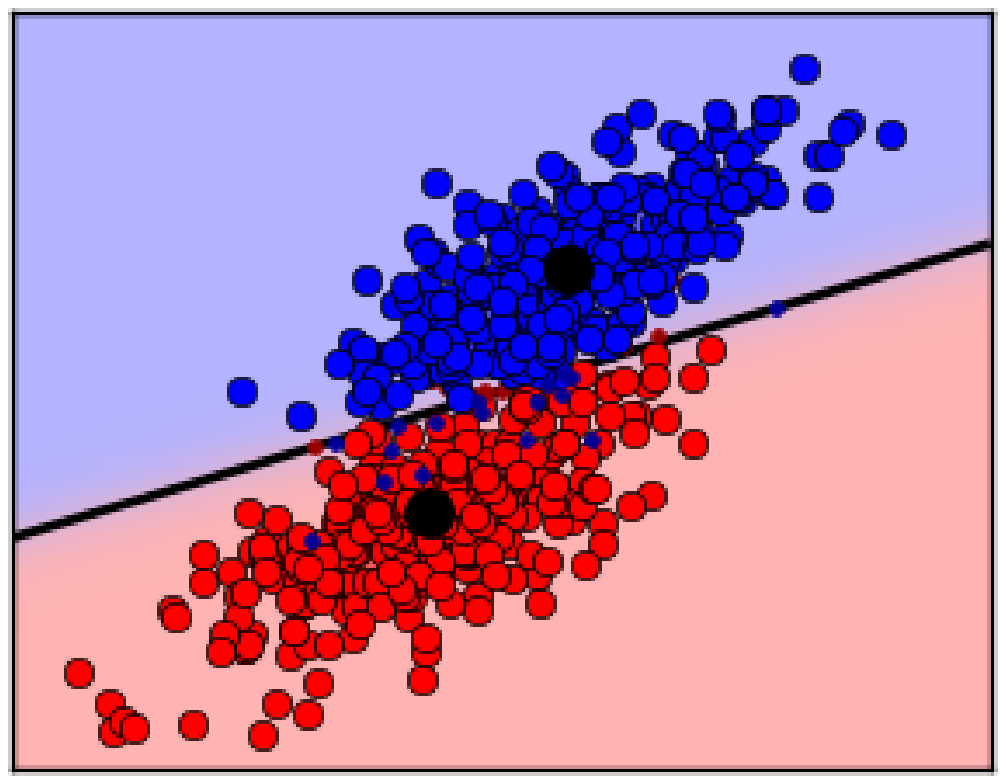}};
  \node[text width=2.5cm,align=center] at (8.85,2.25) {Benign or\\Malignant};
\end{tikzpicture}
	\caption{Illustration of the framework utilized for benign/malignant nodule classification.}
	\label{fig:ch3:framework}
\end{figure*}

The features used to describe a lung nodule can be broadly classified into two: image features 
(geometric, appearance, texture, etc.) and clinical features (age, gender, smocking status, 
medical history, etc.). Focusing on image features, geometric image describe the geometric 
nature of a lung nodule without any reference to the intensity information. 
Several geometric features have been used to characterize a nodule: nodule volume, area, perimeter, diameter, 
surface area, aspect ratio~\cite{Tartar_2014,Lee201043,Ted_Way_2009,armato_2003}.
Some authors have also proposed geometric features that describe the nature of a nodule:
solidity, eccentricity, compactness, circularity, and sphericity~\cite{Tartar_2014,armato_2003}.
Though geometric features are useful for benign-malignant discrimination,
they are rarely used alone and are mostly combined with other image or clinical features (see Table~\ref{tbl:soa:feat_summ}).

Appearance based image features are obtained based on the 
pixel intensity information available on lung CT image. Except gradient features,
they are obtained by looking at each nodule pixel independently (with minimal neighborhood
information). The most widely used appearance image features are gray level region statistics (mean, standard deviation)~\cite{armato_2003},
gray scale histogram (or statistics derived from it)~\cite{Li2005357,armato_2003,Aoy02}, and gradient image features~\cite{Ted_Way_2009}.
These features are very easy to compute and do indeed provide discriminatory information
that arises from intensity difference of benign and malignant nodule due to different tissues.
On the other hand, texture image features, contrary to 
appearance image features, are extracted by analyzing a pixel and its neighborhood
for different patterns. They can be used to characterize shape smoothness, irregularity,
and patterns. Examples include, Fourier descriptors~\cite{Lee201043}, fractal patterns~\cite{Lee201043},
and wavelet descriptors (extracted using wavelet transform)~\cite{MaderoOrozco2015}.
Furthermore, it has also been shown that adding clinical features, if available and when registered without
error, improves performance~\cite{Lee201043,Tartar_2014}. A recent work 
presented by Kumar et al.~\cite{kumar2015lung} has shown that good performance
can be obtained by using automatically identified image features
based on deep learning approach.

On the other hand, classifiers also play an important role in benign-malignant lung nodule classification
as they make the final decision -- classification label. The classifier used
should generalize as much as possible using the data provided in the training stage
so that it can perform well in unseen (test) instances. Several supervised 
linear and non-linear classifiers have been used for lung nodule classification:
Linear examples include, Linear Discriminant Analysis (LDA)~\cite{Li2005357,armato_2003} and
linear Support Vector Machines (SVM)~\cite{MaderoOrozco2015}; 
Non-linear classifiers include, ensemble classifiers (AdaBoost and random forest)~\cite{Tartar_2014},
Artificial Neural Networks (ANN)~\cite{Aoy02}, and Decision Trees~\cite{kumar2015lung}.

\section{ADOPTED CLASSIFICATION FRAMEWORK} \label{sec:system_overview}

In this work, a classical two stage approach, shown in Fig.~\ref{fig:ch3:framework},
 to identify malignant and benign lung nodules from a given lung CT image containing a lung nodule is adopted. 

This framework can be described in three steps: 
\begin{enumerate}
\item Given a lung CT slice with radiologist annotated nodule
margins, crop a rectangular region encapsulating the nodule region of interest (ROI);
\item Extract geometric and appearance image features that characterize the nodule image; and
\item Use a trained discriminatory binary (two class) classifier to label the extracted
feature, hence the nodule, as benign or malignant.
\end{enumerate}

The classifier is trained a priori using labeled positive (malignant) and negative (benign) 
image features extracted from lung nodule dataset with diagnosis information in a supervised manner.
The set of image features and discriminant classifiers used are described in
Sections~\ref{ch3:sec:features} and~\ref{sec:classifiers} respectively. As image
features, a heterogeneous feature composed of three different feature sets: geometric features (nodule diameter, aspect ratio, area, and perimeter),
gray scale histograms, and oriented gradient histograms, is proposed. For the classification task,
five different linear and non-linear classifiers types -- linear (logistic regression, linear support vector machine),
and non-linear (K-nearest neighbor, discrete AdaBoost, and random forest) -- are utilized. 

\section{Image Features} \label{ch3:sec:features}
Image features that capture important cues of a class of data are vital for successful classification
tasks. The proposed feature is composed of the three feature sets described below.

\subsection{Geometrical Features} \label{ch3:ss:gfeats}
Geometrical properties of lung nodules are very important in benign and malignant lung nodule identification,
for example, the larger the size of a nodule, the more likely it is to be malignant~\cite{wahidi_2007}.
Accordingly, four set of geometric features in metric units are extracted from a given annotated lung nodule:
Nodule diameter ($\delta$), Nodule aspect ratio ($\rho$), Nodule area ($a$), and Approximate nodule perimeter ($p$).

\begin{figure}[!ht]  
\vspace{0.25cm}
\centering
\begin{tikzpicture}[every text node part/.style={align=left},xscale=1.25,yscale=1.25]
\small
  \node[inner sep=0pt] (nodule) at (0,0) {\includegraphics[width=.18\textwidth]{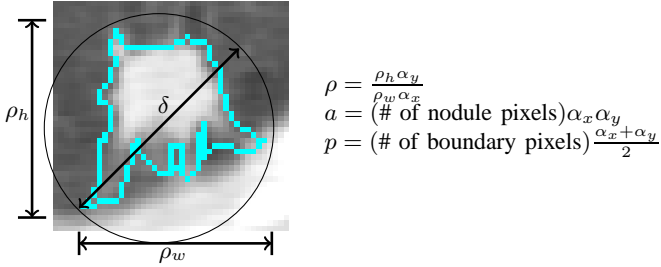}};
  \draw (-0.15,-0.15) circle (1.22cm);
  \draw [line width=0.35mm,-] (-1.35,-1.1) -- (-1.65,-1.1);
  \draw [line width=0.35mm,-] (-1.35,1.0) -- (-1.65,1.0);
  \draw [line width=0.35mm,-] (-1.0,-1.25) -- (-1.0,-1.5);
  \draw [line width=0.35mm,-] (1.05,-1.25) -- (1.05,-1.5);

  \draw [line width=0.35mm,<->] (1.05,-1.375) -- (-1.0,-1.375);
  \draw [line width=0.35mm,<->] (-1.5,-1.1) -- (-1.5,1.0);

  \draw [line width=0.35mm,<->] (-1,-1) -- (0.7,0.7);
  \node[inner sep=0pt] (diam) at (-0.1,0.1)  {$\delta$};
  \node[inner sep=0pt] (aw) at (-1.65,0)  {$\rho_h$};
  \node[inner sep=0pt] (aw) at (0,-1.5)  {$\rho_w$};


  \node[inner sep=0pt] (aw) at (3.4,0)  {$\rho = \frac{\rho_h\alpha_y}{\rho_w\alpha_x}$\\ $a=(\text{\# of nodule pixels})\alpha_x\alpha_y$\\$p=(\text{\# of boundary pixels})\frac{\alpha_x+\alpha_y}{2}$};
\end{tikzpicture}
	\caption{Illustration of the geometrical features extracted from annotated lung nodule CT.}
	\label{fig:ch3:geom_feats}
\end{figure}

Figure~\ref{fig:ch3:geom_feats} visually illustrates the above listed geometric features. The $\alpha_x$ and $\alpha_y$
values are pixel to metric conversion factors along horizontal and vertical axis of the lung CT dicom image.
The nodule perimeter $p$ is described as approximate because an average conversion factor ($\frac{\alpha_x+\alpha_y}{2}$)
is used to convert the nodule boundary provided by a radiologist in pixel to metric unit.  
These geometric features are then used to define a feature vector $\vx_{g} = [\delta,\rho,a,p]^{\mathtt{T}}$.
If the nodule ROI annotation comes from several radiologists (as in the case of LIDC-IDRI lung CT dataset
presented in Section~\ref{subsec:dataset}), the union of all ROIs is considered as the nodule region.

\subsection{Gray Scale Histogram} \label{ch3:ss:gfeats}

The second set of feature considered is lung nodule image gray scale information. To capture the pixel 
appearance information of an imaged object in a rotation and scale invariant manner, gray scale histogram, 
also called intensity histogram, is utilized.

\begin{figure}[!ht]  
\centering
\includegraphics[width=.1\textwidth]{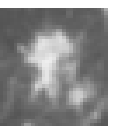}
\includegraphics[width=.18\textwidth]{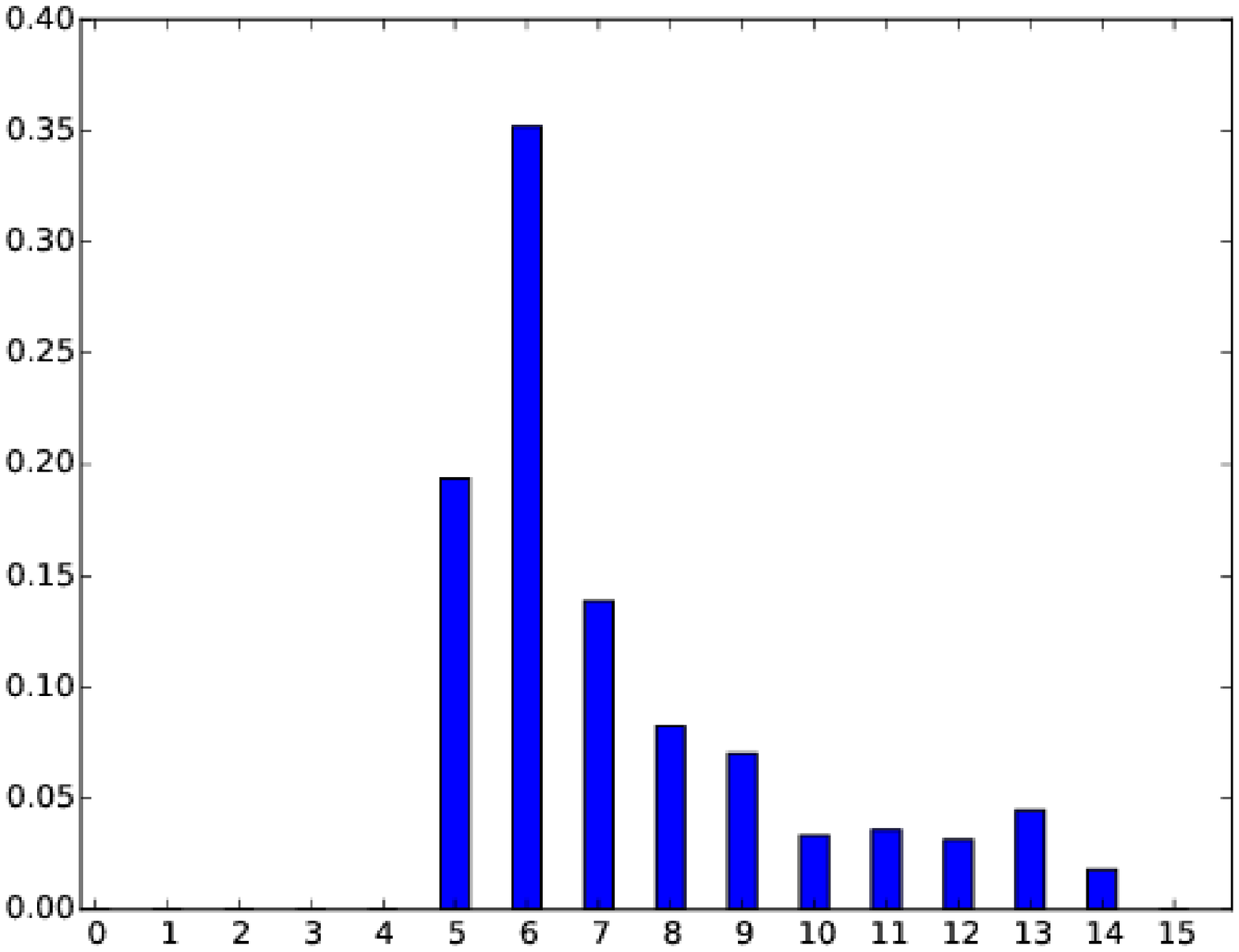}
\includegraphics[width=.18\textwidth]{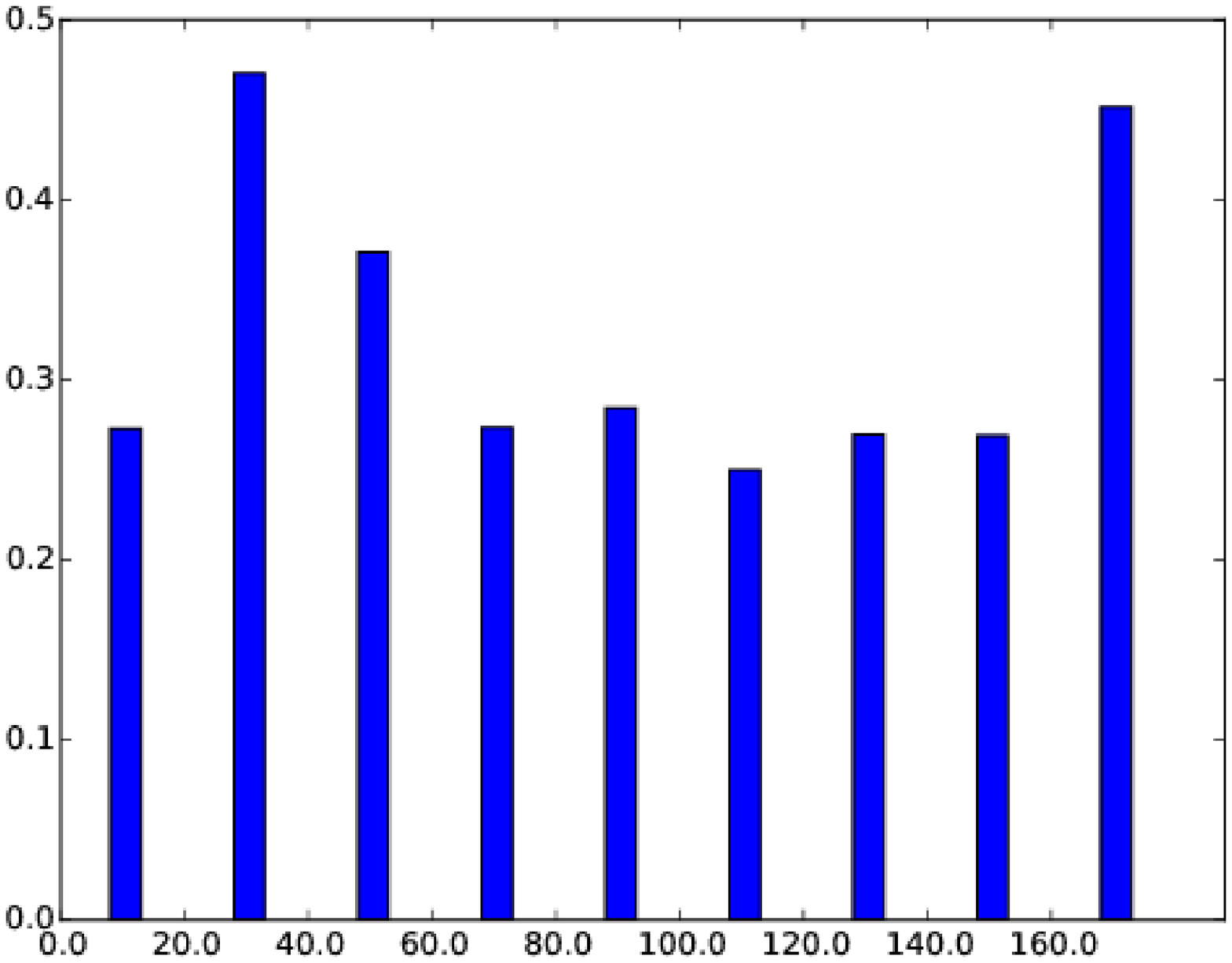}\\
\subfloat[\label{fig:ch3:hist_feature_a}]{\includegraphics[width=.1\textwidth]{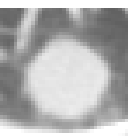}}
\subfloat[\label{fig:ch3:hist_feature_b}]{\includegraphics[width=.18\textwidth]{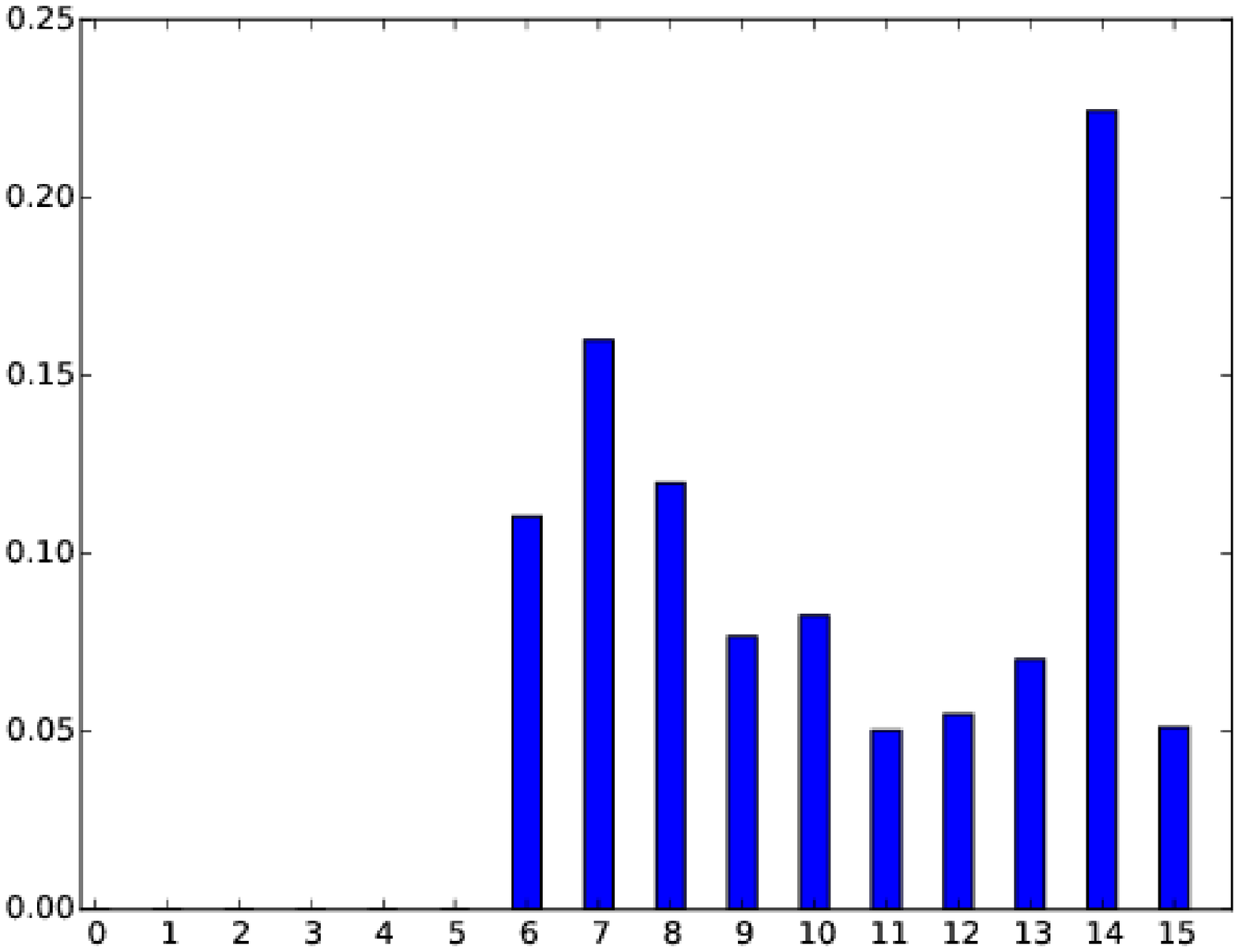}}
\subfloat[\label{fig:ch3:hist_feature_c}]{\includegraphics[width=.18\textwidth]{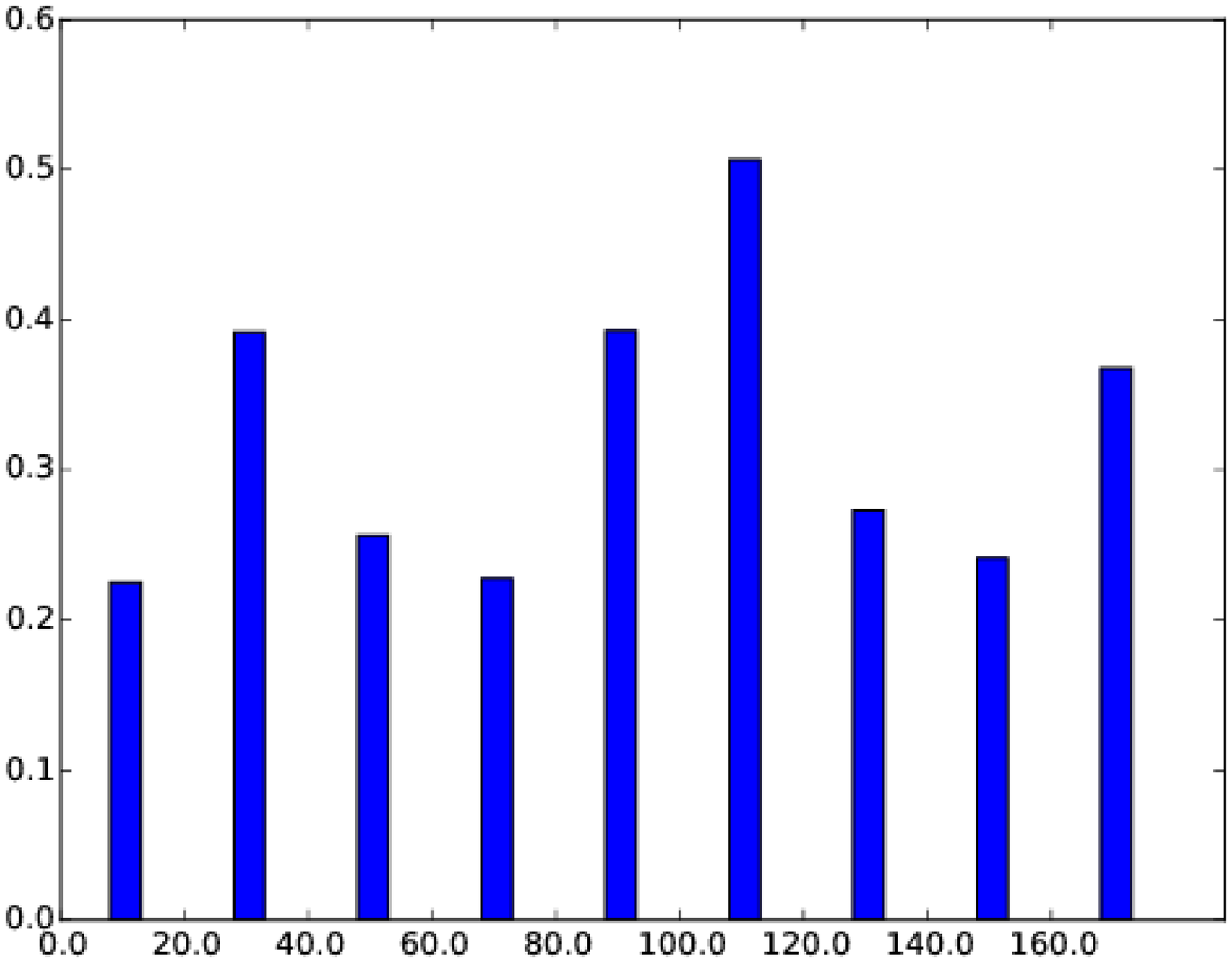}}
	\caption{(a) Lung nodules, (b) Gray scale histogram (16 bins), (c) Histogram of oriented gradient. Top row is of a benign nodule, bottom is that of a malignant one.}
	\label{fig:ch3:hist_feature}
\end{figure}

Given an intensity lung nodule image, as in Fig.~\ref{fig:ch3:hist_feature_a}, a gray scale histogram of $B$ bins is extracted
by first dividing the image range in $B$ equally spaced gray scale value ranges. Then for each pixel
value, the corresponding bin value is incremented by one. Finally, the obtained histogram is normalized.
Here, a gray scale histogram of $16$ bins is used (i.e., $B = 16$). Sample histograms of a benign
and malignant nodules are shown in Fig.~\ref{fig:ch3:hist_feature_b} top and bottom respectively.

\subsection{Oriented Gradient Histogram} \label{ch3:ss:hog}

The third feature type considered is image gradient histogram. The gradient information
in an image provides a lot of information about the nature of the object presented in the image. 
Gradient magnitude and orientation based features are the most discriminant and most
successfully used features in object detection and classification tasks~\cite{Dollar2014}.
The oriented gradient histogram is computed first by
determining the image gradient (magnitude and orientation) at each pixel of the given image
containing a lung nodule. Then a histogram whose bins represent gradient orientations is constructed
by adding the gradient magnitude of the pixel at the corresponding histogram bin. Basically, the horizontal
axis of the histogram corresponds to gradient orientation and the vertical axis corresponds to the binned
gradient magnitude. Contrary to most approaches in the literature that concatenate localized histograms to keep
spatial information, one global histogram per image is computed in this work to minimize its variance to
image (or lung nodule) rotation. A contrast insensitive (considering only $[0^o,180^o]$ magnitude
orientation) oriented gradient histogram of $9$ bins is used. Sample illustrative histograms 
are shown in Fig.~\ref{fig:ch3:hist_feature_c}.

Finally, all the three extracted feature sets, geometric, gray scale histogram,
and oriented gradient histogram, are combined to create a $29$ dimensional heterogeneous feature set (denoted with $\vx$).

\section{CLASSIFIERS} \label{sec:classifiers}

Five commonly used linear and non-linear
classifiers are investigated. The linear ones consist of Logistic Regression classifier and
Linear Support Vector Machine. The non-linear ones include K-Nearest Neighbor, Discrete AdaBoost,
and Random Forest classifiers. All the classifiers considered are supervised classifiers, which
are trained using labeled positive and negative training data (for two class classification problem as in
this work). Given a labeled set of $n$ training instances, $\left\{(\vx_i,y_i)\right\}^n_{i=1}$
with $y_i\in\{-1,+1\}$ and $\vx\in\R{}^{d}$ (a $d$ dimensional feature vector), the classifier
learns a classification rule $f: \vx\to{}y$, that maps the feature vector $\vx$ to its label $y$.
The classifiers also have a function $h(\vx)$ that provides a continuous score
that acts as a confidence indicator of positive label. In fact, the classification rule $f$ is derived
from $h$ by thresholding the score with a tuned (learned) threshold value $\theta$: if
the score is equal to or above $\theta$, a positive label (malignant)
is assigned, and otherwise a negative label (benign) is assigned.

Each classifier's hyper-parameters are tuned via cross-validation. This includes, the $C$
of Logistic Regression and Linear SVM, the $K$ of K-NN, decision tree depth $D$ of AdaBoost,
and the number $N$ and depth $D$ of decision trees used in the Random Forest classifier
(these parameters are defined according to~\cite{scikit-learn}).

\section{EXPERIMENTS AND RESULTS} \label{sec:experiments_results}

\subsection{Evaluation Metrics} \label{subsec:metrics}
This work deals with a binary classification task. To quantitatively evaluate a trained classifier operating
on a fixed point (once best case classifier thresholds $\theta$ rules have been identified via
cross validation), the following standard measures are used~\cite{Bowen2014}:
\begin{eqnarray}
  Sensitivity: {S}_e = \frac{TP}{TP + FN} \\
  Specificity: {S}_p = \frac{TN}{TN + FP} \\
  Accuracy: {A} = \frac{TN + TP}{TP+TN+FN+FP} \\
  F-measure = 2*\left(\frac{{S}_e{S}_p}{{S}_e+{S}_p}\right)
\end{eqnarray}

True Positive ($TP$), False Negative ($FN$), True Negative ($TN$), and False Positive ($FP$) 
are defined in the obvious sense. Sensitivity characterizes
how well the classifier correctly recognizes malignant nodules, and specificity that of benign nodules.
Accuracy measures the proportion of total data correctly classified. The F-measure, contrary to the common
formulation based on Precision-Recall, is defined here as the harmonic mean of sensitivity and specificity
to provide a single measure that combines both. Sensitivity-specificity ROC curve is used to 
characterize classifier performance over several operating points. It is then summarized by obtaining
 the Area Under the Curve (AUC).

\subsection{Dataset} \label{subsec:dataset}
The proposed benign-malignant classification framework is primarily
trained and tested using the publicly available LIDC-IDRI lung CT image
dataset~\cite{lidc_dataset}. This dataset is of particular interest in this work because it provides
diagnosis data for a subset of the subjects -- we use the data from $95$ subjects for which
accurate diagnostic label could be established. 
The nodule level diagnosis is marked as: 0 - unknown, 1 - benign, 2 - malignant (primary lung cancer),
and 3 - malignant (metastatic). Furthermore, nodules with only benign and malignant labels (1,2,3) are considered which
further reduced the data (all nodules with a label 0, for unknown, are not considered).
This resulted in $52$ subjects with malignant nodules and $21$ subjects with benign nodules
with a total of $458$ and $107$ individual lung CT slice nodules respectively (see Table~\ref{table:ch4:diagnosis}). Out of
this $65\%$ are used for training and the rest are used for testing.

\begin{table}[!ht]
\resizebox{0.49\textwidth}{!}{
\centering
\begin{tabular}{cccc}
\toprule
				&	unknown (0)		&	benign (1)		&	malignant (2 and 3)	\\
\toprule
\# of subjects	&	22				&	21				&	52		\\
\# of CT slices	&	74				&	107				&	458		\\ \hline
\# of CT slices (train / test)	&	--	& 66 / 41			&	301 / 157	\\	
\bottomrule
\end{tabular}
}
\caption{LIDC-IDRI dataset summary of nodule diagnosis information.}
\label{table:ch4:diagnosis}
\end{table}

\subsection{Implementation Details} \label{subsec:implementation}
The lung nodule benign-malignant classification framework presented in this work has been completely implemented
in python. The dicom data obtained from the LIDC-IDRI dataset is normalized to $[0,256)$
discrete values. When extracting histogram features (gray scale and oriented gradient), a rectangular region
encompassing the union of all radiologist nodule boundary annotation with an additional $5\%$ margin to include
background information is used. All described features are mean and variance normalized to approximately 
follow a normally distributed data. This is a common requirement
for the classification algorithms used which are based on scikit-learn~\cite{scikit-learn}.
Given the small number of training/test data available, a 5-fold cross validation setup is
used to determine model variables. Once the suitable variable is identified, the classifier 
is retrained using the entire training data. Finally, this trained classifier is evaluated on 
the test set to provide definitive evaluation metrics, both operating point metrics and ROC curve.

\subsection{Results} \label{subsec:results}

\begin{figure}[!ht]
\centering
\begin{tikzpicture}[every text node part/.style={align=center},xscale=1.75,yscale=1.75]
\tikzstyle{block} = [inner sep=2mm, draw]
\small
\end{tikzpicture}
\includegraphics[width=0.4\textwidth]{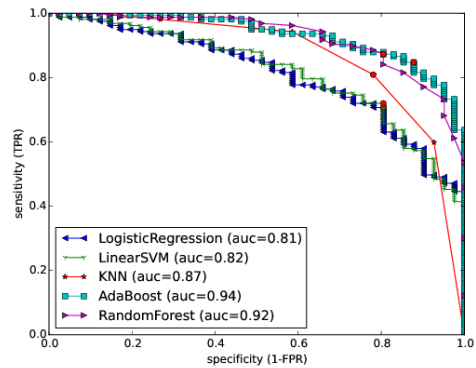}
\caption{Test set ROC curves obtained using the combined heterogeneous features.}
\label{fig:ch4:test_comb}
\end{figure}

The final test results obtained using the combined heterogeneous feature set
with the different classifiers, test ROC curves,  are shown in Fig.~\ref{fig:ch4:test_comb}.
The AUC and operating point evaluation results are also detailed in Table~\ref{tbl:ch4:test_comb}.
Overall, the best AUC result is obtained by AdaBoost and is $0.94$ which is very close to
a perfect score.  This AdaBoost classifier also achieves the best specificity, accuracy, 
and F-measure. The second best results are obtained using the random forest classifier. 
The results obtained by the non-linear classifiers are much better than the linear classifier cases.

\begin{table}[!ht]
\centering
\resizebox{0.485\textwidth}{!}{
\begin{tabular}{cc|ccccccc}
\toprule
Classifier 			&  Parameter(s) &	 AUC	&  Sensitivity	&	Specificity	&	Accuracy	& 	F-measure	\\
\toprule
Logistic Regression &  $C=2.0$  	&	0.81	&   0.71	&	0.80		&	0.73		&	0.75		\\
Linear SVM			&  $C=0.25$  	&	0.82	&   0.72	&	0.80		&	0.74		&	0.76		\\
K-NN					&  $K=5$  		&	0.87	&   0.81	&	0.78		&	0.80		&	0.79		\\
AdaBoost			&  $D=5$  		&\best{0.94}&	0.82	&\best{0.93}	&\best{0.84}	&\best{0.87}	\\
Random Forest		&  $D=25$,$N=40$&	0.92	&   0.80		&	0.90		&	0.82		&	0.85		\\ \hline
Kumar et al.~\cite{kumar2015lung}	&	--			&	--		&	\best{0.83}		&	0.21		&	0.75		&	0.34			\\
Kumar et al.~\cite{kumar2015arxiv}	&	--			&	--		&	0.79		&	0.76		&	0.78		&		0.77		\\
\bottomrule
\end{tabular}
}
\caption{Test set results obtained using the combined heterogeneous feature set with optimized classifiers
and comparisons with the state-of-the-art.
\best{Best results} on each metric are highlighted.}
\label{tbl:ch4:test_comb}
\end{table}

These results are very promising. Unfortunately, due to the nature of the data used,
a direct comparison with results in the literature is not valid. As described, the LIDC-IDRI
diagnosis data does not provide an absolute position of the referenced nodule. 
This means that it is only possible to use nodule data in certainty \emph{if and only if} a patient has 
only one identified lung nodule (which reduced the total number of subjects to use from 157 to 95).  
For the sake of comparison, the last two rows of Table~\ref{tbl:ch4:test_comb} report state-of-the-art
results in the literature obtained using the LIDC-IDRI dataset. Except the  sensitivity of $83\%$,
our best approach based on AdaBoost and the proposed heterogeneous features outperforms their reported results.
We obtain a $9\%$ and $6\%$ improved accuracy compared to~\cite{kumar2015lung} and~\cite{kumar2015arxiv} respectively.
We also obtain a significantly improved F-measure.

\section{CONCLUSIONS AND FUTURE WORKS} \label{sec:conclusion}

This work investigated an automated framework for lung nodule benign-malignant
classification based on lung CT images with annotated nodules. 
The experimental results presented in this work make it possible to make
the following three conclusive observations based on the dataset utilized:
(i) Image features provide useful cues that are useful for benign-malignant
lung nodule classification, 
(ii) Heterogeneous features lead to improved classification accuracy,
compared to the constituent counterparts, as they combine various complementary cues,
and (iii) In general non-linear classifiers, especially ensemble classifiers,
 are better suited for lung nodule benign-malignant classification.
The experimental results on the LIDC-IDRI public dataset are very encouraging
correctly classifying $82\%$ of malignant and $93\%$ of benign nodules
on unseen test data at best. 

Possible future lines of investigations include: addition of 
texture image features, e.g., Local Binary Patterns (LBP),
consideration of volume CT image features,  and probabilistic data
fusion strategies to incorporate clinical features at a higher reasoning
level.



\bibliographystyle{plain}
\bibliography{02-MainWriting}

\end{document}